\documentclass[letterpaper]{article} 
\usepackage{aaai25}  
\usepackage{times}  
\usepackage{helvet}  
\usepackage{courier}  
\usepackage[hyphens]{url}  
\usepackage{graphicx} 
\usepackage{natbib}  
\usepackage{caption} 
\usepackage{algorithm}

\usepackage{newfloat}
\usepackage{listings}
\DeclareCaptionStyle{ruled}{labelfont=normalfont,labelsep=colon,strut=off} 
\floatstyle{ruled}
\newfloat{listing}{tb}{lst}{}
\floatname{listing}{Listing}
\usepackage{booktabs}
\usepackage{xcolor}
\usepackage{url}
\usepackage{tabularx}
\usepackage{graphicx}
\usepackage{adjustbox}
\usepackage{amsmath}
\usepackage{algorithm}
\usepackage{algpseudocode}
\usepackage{cite}
\usepackage{siunitx}
\begin{document}
\title{Look Inside for More: Internal Spatial Modality Perception for 3D Anomaly Detection}

\author{
    Hanzhe Liang\textsuperscript{\rm 1, 2},
    Guoyang Xie\textsuperscript{\rm 4},
    Chengbin Hou\textsuperscript{\rm 5},
    Bingshu Wang\textsuperscript{\rm 6},
    Can Gao\textsuperscript{\rm 1}\thanks{Corresponding authors.},
    Jinbao Wang\textsuperscript{\rm 3,7}$^*$\\
}

\affiliations{
    \textsuperscript{\rm 1} College of Computer Science and Software Engineering, Shenzhen University, Shenzhen, China.\\     
    \textsuperscript{\rm 2} Shenzhen Audencia Financial Technology Institute, Shenzhen University, Shenzhen, China.\\
    \textsuperscript{\rm 3} National Engineering Laboratory for Big Data System Computing Technology, Shenzhen University, Shenzhen, China.\\
    \textsuperscript{\rm 4} Department of Intelligent Manufacturing, CATL, Ningde, China.\\ 
    \textsuperscript{\rm 5}	School of Computing and Artificial Intelligence, Fuyao University of Science and Technology, Fuzhou, China.\\
    \textsuperscript{\rm 6} School of Software, Northwestern Polytechnical University, Xi’an, China.\\
    \textsuperscript{\rm 7} Guangdong Provincial Key Laboratory of Intelligent Information Processing, Shenzhen, China.\\
    
    2023362051@email.szu.edu.cn, guoyang.xie@ieee.org, chengbin.hou10@foxmail.com, wangbingshu@nwpu.edu.cn, davidgao@szu.edu.cn, wangjb@szu.edu.cn\\
}

 \maketitle

\begin{abstract}
3D anomaly detection has recently become a significant focus in computer vision. Several advanced methods have achieved satisfying anomaly detection performance. However, they typically concentrate on the external structure of 3D samples and struggle to leverage the internal information embedded within samples. Inspired by the basic intuition of why not look inside for more, we introduce a straightforward method named Internal Spatial Modality Perception~(ISMP) to explore the feature representation from internal views fully. Specifically, our proposed ISMP consists of a critical perception module, Spatial Insight Engine~(SIE), which abstracts complex internal information of point clouds into essential global features. Besides, to better align structural information with point data, we propose an enhanced key point feature extraction module for amplifying spatial structure feature representation. Simultaneously, a novel feature filtering module is incorporated to reduce noise and redundant features for further aligning precise spatial structure. Extensive experiments validate the effectiveness of our proposed method, achieving object-level and pixel-level AUROC improvements of 3.2\% and 13.1\%, respectively, on the Real3D-AD benchmarks. Note that the strong generalization ability of SIE has been theoretically proven and is verified in both classification and segmentation tasks.
\end{abstract}

\section{Introduction}
3D anomaly detection (AD) plays a crucial role in industrial and medical applications by identifying abnormalities in complex structures. Traditional methods, such as BTF~\cite{horwitz2022featureclassical3dfeatures}, primarily focus on single-sample analysis, while recent deep learning-based approaches have improved detection by incorporating cross-sample information. However, these methods often rely on intuitive feature extraction, which may overlook deeper anomalies. Researchers are exploring different strategies to uncover finer details, with some emphasizing 3D data alone and others integrating multi-modal approaches.

\begin{figure}[t]
    \centering
    \includegraphics[width=1\linewidth]{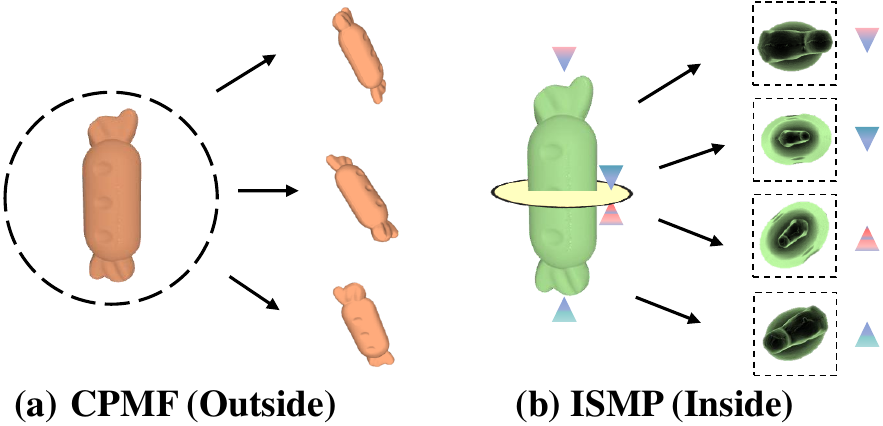}
    \caption{Visualization of internal and external perception. Compared with external view (CPMF), our method (ISMP) projects from the internal view, better capturing the different-shaped protrusions in the 3D structure.}
    \label{fig:first}
\end{figure}

The methods centered on 3D structures emphasize the unique feature representation of the structure. For example,~\cite{bergmann2022anomalydetection3dpoint} used geometric descriptors with a teacher-student model to achieve promising results, while~\cite{rudolph2022asymmetricstudentteachernetworksindustrial} introduced asymmetric networks to enhance discrimination further. Additionally,~\cite{li2023scalable3danomalydetection} focused on improving local feature representations, and~\cite{kruse2024splatposedetectposeagnostic} proposed leveraging pose information for better anomaly detection across different viewpoints. Despite these advancements, many methods start from an intuitive structure, potentially leading to incomplete information coverage.
On the other hand, multi-modal methods provide richer feature representations by integrating different data modalities. For instance, combining RGB 2D and 3D data~\cite{wang2023multimodalindustrialanomalydetection} or using independent evaluations of both~\cite{pmlr-v202-chu23b} has enhanced detection capabilities.~\cite{zavrtanik2023cheatingdepthenhancing3d} leveraged depth and RGB information to identify anomalies better, while~\cite{bhunia2024looking3danomalydetection} advanced 2D-3D detection by constructing a query image database. However, challenges such as feature alignment losses and increased sensor data demands persist. To address these issues,~\cite{cao2023complementarypseudomultimodalfeature} introduced a pseudo-modal approach that projects 3D data into 2D images for supplementary information. While this method offers a more comprehensive representation, it still neglects internal structural details, resulting in incomplete feature coverage and reducing detection performance.

\textbf{Could we shift focus toward internal information for more comprehensive anomaly detection?}
To tackle the challenges of insufficient internal information utilization and difficulties in aligning data across different modalities in anomaly detection using pseudo-modalities, we propose a novel method centered on internal spatial pseudo-modalities. Figure 1 shows the comparison between our method's internal perception and counterparts that obtain global features from the outside. Our approach effectively captures the internal characteristics of 3D structures, even in low-sample environments, by leveraging the internal spatial features of point clouds. It facilitates better interactions between internal structures and surface regions, creating a complementary relationship between internal and external information. The core of our method, the Internal Spatial Modality Perception (ISMP) framework, includes a Spatial Insight Engine (SIE) that captures global features, an enhanced feature extraction module for local details, and a feature filtering module to suppress redundant data. Together, these components significantly improve anomaly detection accuracy. Note that our SIE indicates strong generalization capabilities, making it suitable for a broader range of point cloud tasks.

The main contributions of this paper are summarized as follows:
\begin{itemize}
    \item To our best of knowledge, we are the first to focus on the internal structure of point clouds, thereby improving the extraction of internal structural features. 
    \item A new Internal Spatial Modality Perception (ISMP) module and an Enhanced feature extraction combined with a feature filtering module, are designed to improve the perception and alignment of local features of key points.
    \item The feasibility of Spatial Insight Engine (SIE) is explored in tasks of classification and segmentation, emphasizing the strong generalization ability of the internal spatial pseudo-modality.
    \item Numerous experiments demonstrate the superiority of ISMP, surpassing the state-of-the-art methods on Real3D-AD with 13.1\% and 3.2\% improvements in P-AUROC and O-AUROC.
\end{itemize}

\subsection{2D Anomaly Detection}
2D image anomaly detection, a widely studied area, typically involves two main components: feature extraction and feature modeling. Feature extraction aims to derive discriminative features that distinguish normal from anomalous data. In contrast, feature modeling captures the distribution of normal features and detects deviations when anomalies are present. Early methods focused on learning features from scratch, such as through autoencoders and inpainting tasks, with notable approaches like RIAD, and DRAEM making significant strides in this area~\cite{Bergmann_2019, park2023excisionrecoveryvisualdefect, zavrtanik2021draemdiscriminativelytrained}. However, recent advancements have demonstrated the effectiveness of using pre-trained networks for anomaly detection. Techniques like knowledge distillation, as employed in ST and AST, align features between teacher and student networks to detect anomalies, addressing issues like overgeneralization~\cite{yamada2022reconstructionstudentattentionstudentteacher, rudolph2022asymmetricstudentteachernetworksindustrial}. Further innovations include normalizing flow and memory bank techniques to model normal feature distributions more effectively~\cite{gudovskiy2021cflowadrealtimeunsupervisedanomaly}. These developments improve 2D anomaly detection and lay the groundwork for extending these methods to 3D and multi-modal detection, driving further progress in the field.

\begin{figure*}[t] 
    \centering 
    \includegraphics[width=1\linewidth]{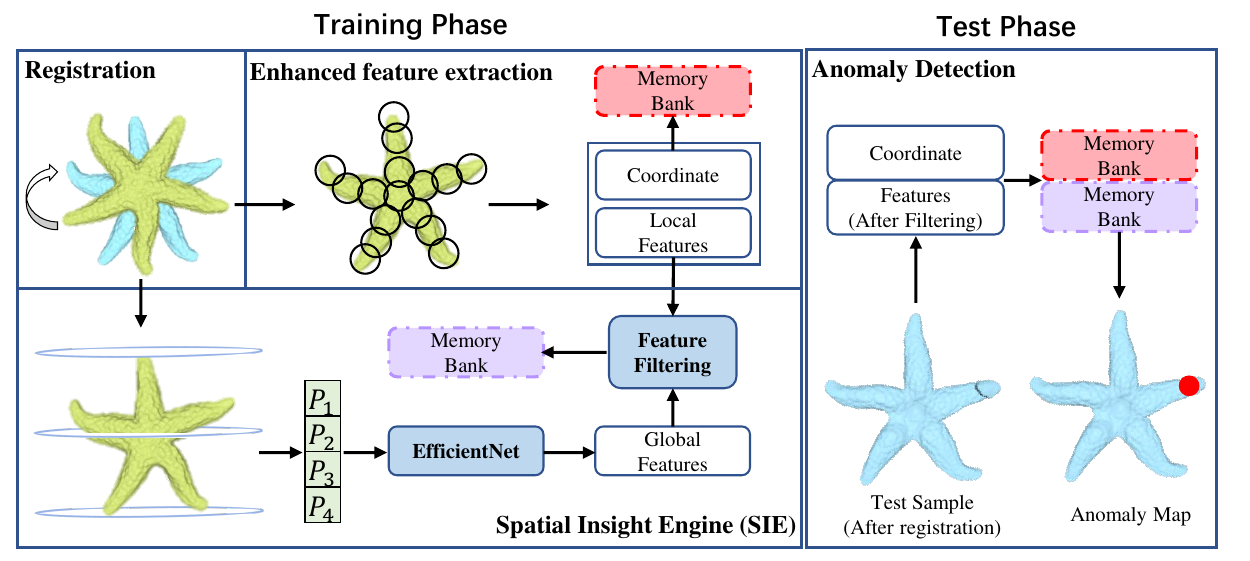} 
    \caption{Overview of our method. We start by matching the point cloud according to RANSAC~\cite{9318535}. During the training phase, we create feature and coordinate memory banks, using enhanced feature extraction to capture local information and constructing a local feature matrix. We extract global features using the SIE and align them with the local ones. $P_{1}$, $P_{2}$, $P_{3}$, and $P_{4}$ are the four projection slices extracted respectively. Then, we employ a feature filtering module to suppress redundant information, resulting in the final feature matrix. In the inference phase, we obtain the final feature matrix and compute the nearest neighbors in the memory bank. 
    Finally, we input the coordinates into the coordinate memory bank to find the closest regular sample coordinates, calculating the final score of the sample points based on both memory banks.} 
    \label{model} 
\end{figure*}

\subsection{3D Anomaly Detection}
3D anomaly detection, such as point cloud AD, is crucial in domains like autonomous vehicular navigation and industrial inspections ~\cite{solaas2024systematicreviewanomalydetection, Cui_2022}. Deep learning-based anomaly detection leverages neural networks to capture intricate point cloud structures. Techniques like PatchCore and its successors have made significant strides by learning point cloud representations directly from raw data ~\cite{tu2024selfsupervisedfeatureadaptation3d,roth2022totalrecallindustrialanomaly}. These methods emphasize efficient feature extraction and fusion, crucial for effective anomaly detection. Advanced approaches like PointNet++ and Point Transformer improve feature extraction by incorporating hierarchical and attention mechanisms ~\cite{wu2024pointtransformerv3simpler, zhao2021pointtransformer, qi2017pointnetdeephierarchicalfeature}. Additionally, techniques like PointMAE and PointMLP further enhance local feature extraction and fusion ~\cite{pang2022maskedautoencoderspointcloud, ma2022rethinkingnetworkdesignlocal}. Mathematical strategies, including coupled Laplacian eigenmaps and locality-sensitive methods, also contribute to more nuanced point cloud representations, enhancing 3D anomaly detection ~\cite{bastico2024coupledlaplacianeigenmapslocallyaware, chen2023trajectoryformer, bergmann2022anomalydetection3dpoint}. Finally, point cloud coordinates, exemplified by methods like FPFH, provide essential feature information for anomaly detection ~\cite{5152473}.

In parallel, pseudo-modality techniques, which simulate modality data through a single modality or fabricated features, aim to enhance feature representation by combining diverse types of information. Recent advancements have addressed some of these shortcomings. For instance, methods discussed in~\cite{cao2023complementarypseudomultimodalfeature} have focused on leveraging pseudo-modal features from multiple viewpoints. Building on this,~\cite{bhunia2024looking3danomalydetection} further improves the capture of texture information by aligning and transposing 2D features onto 3D point clouds using extensive 2D image databases for referencing. However, despite these advancements, these methodologies commonly need to work on fully exploiting the intricate internal structure of point clouds. They predominantly focus on extrinsic information while overlooking essential internal complexities. Consequently, this oversight limits their effectiveness in fully capturing the nuanced internal features necessary for comprehensive anomaly detection.

\section{Method}

\subsection{Spatial Insight Engine}
\begin{figure}[ht]
    \centering
    \includegraphics[width=0.85\linewidth]{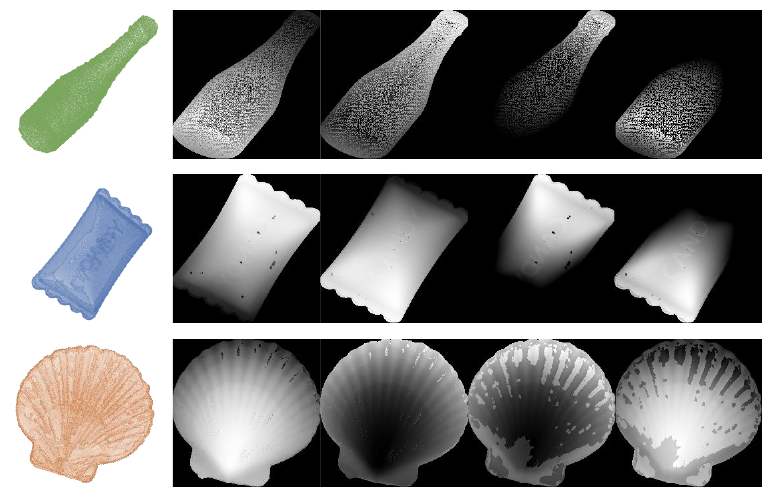}
    \caption{Visualization of projection slices. The images are the original image, $P_1$, $P_2$, $P_3$, and $P_4$, respectively.}
    \label{fig:proj_slices}
\end{figure}

Figure~\ref{model} provides an overview of our method. We have developed a robust model (SIE) that seamlessly converts 3D point cloud structures into 2D pseudo-modal data using advanced four projection slices ($P_1$, $P_2$, $P_3$, $P_4$), as shown in Figure~\ref{model}. These meticulously designed slices conduct thorough top-down and bottom-up analyses at the point cloud's top, middle, and bottom. More notably, the middle slice ($P_2$, $P_3$) expertly partitions the point cloud into two parts, working synergistically with the other slice to extract comprehensive features from both segments. The visualization of the projection slices is shown in Figure~\ref{fig:proj_slices}.

Taking the upper part of the point cloud as an example, we explain why extracting features from the SIE can capture more information and effectively detect anomalies.

\textbf{Information Capturing Analysis.}
The amount of information can be defined as:
\begin{equation}
\mathbf{P} = \{\mathbf{p}_i = (x_i, y_i, z_i) \mid i \in \{1, 2, \ldots, N\}\},
\end{equation}
where \( \mathbf{P} \) is the set of points, \( \mathbf{p}_i \) represents individual points with coordinates \( (x_i, y_i, z_i) \), and \( N \) is the total number of points.
And the midpoint $z_{mid}$ is defined as:
\begin{equation}
z_{mid} = \frac{z_{\min} + z_{\max}}{2},
\end{equation}
where \( z_{mid} \) is along the z-axis, \( z_{\min} \) and \( z_{\max} \) are the minimum and maximum z-coordinates, respectively.
\begin{equation}
I_{top} = \sum_{i=1}^N (z_{\max} - z_i),
\end{equation}
where \( I_{top} \) is the top information, \( z_i \) is the z-coordinate of point \( i \).
Based on our SIE calculation, we have the global information:
\begin{equation}
I_{global} = \sum_{i=1}^N \left[(z_{\max} - z_i) + \max(0, z_i - z_{mid})\right].
\end{equation}
After rewriting, we have:
\begin{equation}
I_{global} = I_{top} + \sum_{i:z_i \geq z_{mid}} (z_i - z_{mid}) \geq I_{top}.
\end{equation}

Therefore, we observe that $I_{global}$ has more information than $I_{top}$, which is standard external projection manner. The same goes for the lower half of the point cloud. In this way, the final information obtained by the point cloud will be more reliable than if only external modes are used.

\textbf{Anomaly Detection.}
Deep information contains important exception information~\cite{Liu_2024}. An anomaly can be detected when the discrepancy between the depth values from the two views significantly deviates from the expected range for normal points. That is
\begin{equation}
|\Delta D(\mathbf{p}_i) - \mu_{\Delta D}| > k \sigma_{\Delta D},
\end{equation}
where \( \Delta D(\mathbf{p}_i) \) is the discrepancy between top-down and middle-up depth values. Besides, \( \mu_{\Delta D} \) and \( \sigma_{\Delta D} \) are the mean and standard deviation of \( \Delta D \) for average points respectively, with \( k \) as a threshold constant.

Given these constraints, the SIE enhances global information by observing from an internal perspective, significantly improving anomaly detection compared to relying solely on external spatial capture.

\subsection{Enhanced Feature Extraction}
Following the instructions in the relevant work, we utilize Farthest Point Sampling (FPS) to obtain a set of center points, treating the k-nearest neighbors around each center point as a patch for processing~\cite{qi2017pointnetdeephierarchicalfeature}. Following the PointMAE method, we derive the patch features~\cite{pang2022maskedautoencoderspointcloud}. Then, we perform feature extraction on the center points according to FPFH, obtaining more comprehensive features~\cite{5152473}. We have:
\begin{equation}
\text{FPS}(X) = \{x_i\}_{i=1}^m,
\end{equation}
where \(X\) represents the original point cloud and \(\{x_i\}\) are the sampled center points.

For each center point \(x_i\), we define its patch \(P_i\) as:
\begin{equation}
P_i = \{x \in X \mid \|x - x_i\| \leq r \},
\end{equation}
where \(r\) is the radius defining the neighborhood of \(x_i\). Using PointMAE, we extract features for each patch (\(P_i\)). Then, we use Fast Point Feature Histograms (FPFH) to further enhance feature representation for each center point.

\subsection{Feature Filtering Module}
The information extracted from the point cloud is often too miscellaneous, and we usually need to perform noise reduction and other processing on the direct information extracted from the point cloud~\cite{cao2023pointjemselfsupervisedpointcloud}. The Laplacian transform~\cite{kipf2017semisupervisedclassificationgraphconvolutional} is widely used in feature filtering to enhance the quality of features by removing noise and redundant information. It helps in achieving better feature representation and alignment, especially in high-dimensional data such as point clouds~\cite{shao2017attributecompression3dpoint,ghojogh2022laplacianbaseddimensionalityreductionincluding}. By applying the Laplacian transform, models can achieve smoother and more accurate feature extraction, which is crucial for tasks requiring precise geometric representations. 

The Laplacian matrix \(L\) is denoted as:
\begin{equation}
L = D - A,
\end{equation}
where \(D\) is the degree matrix and \(A\) is the adjacency matrix of the graph. This transformation allows for the enhancement of the overall feature quality by smoothing out irregularities and focusing on the intrinsic geometric structure~\cite{zeng20193dpointclouddenoising}.

To achieve better alignment of features from SIE and enhancements, we develop a controllable feature filtering module using the Laplace transform to enhance geometric features in point clouds. This method is outlined in pseudo-code and relies on specific parameters. The process can be summarized by the following equation:
\begin{equation}
\text{Fill}(X| \alpha, \beta, \gamma) = X_{\text{enhanced}},
\end{equation}
where $\alpha$, $\beta$, and $\gamma$ are the parameters that control the influence of the enhanced Laplacian, the decay rate of the weight matrix, and the contribution of the anomaly metric, respectively. $X$ is the original feature matrix, and $X_{\text{enhanced}}$ is the resulting enhanced feature matrix.
The overall implementation of this module is shown in Algorithm~\ref{al1}.

\begin{algorithm}[ht]
\small
\caption{Feature Filtering Module}\label{al1}
\begin{algorithmic}[1]
\Procedure{Filtering}{$X, \alpha, \beta, \gamma$}
    \State $n \gets X.shape()$
    \State $A \gets \mathbf{1}_{n \times n} - \mathbf{I}_n$
    \State $D \gets \text{diag}(A.sum(\text{dim}=1))$
    \State $M \gets \text{cdist}(X, X, p=2)$
    \State $W \gets \exp(-\beta \cdot M)$ \Comment{ Control similarity weighting}
    \State $A' \gets A \cdot W$
    \State $D' \gets \text{diag}(A'.sum(\text{dim}=1))$
    \State $D'^{-0.5} \gets \text{diag}(D'.diag()^{-0.5})$
    \State $L_{\text{sym}}' \gets \mathbf{I}_n - D'^{-0.5} \cdot A' \cdot D'^{-0.5}$
    \State $mask \gets A == 0$
    \State $masked\_M \gets M.masked\_fill(mask, \text{nan})$
    \State $E \gets \text{nanmean}(masked\_M, \text{dim}=1)$
    \State $L_{\text{final}} \gets L_{\text{sym}}' + \gamma \cdot \text{diag}(E)$ \Comment{Adjust Laplacian matrix}
    \State $X_{\text{enhanced}} \gets (\mathbf{I}_n + \alpha \cdot L_{\text{final}}) \cdot X$ \Comment{ Enhance features}
    \State $X_{\text{max}} \gets X.max()$
    \State $X_{\text{enhanced}} \gets X_{\text{enhanced}} \cdot (X_{\text{max}} / X_{\text{enhanced}}.max())$
    \State \Return $X_{\text{enhanced}}$
\EndProcedure
\end{algorithmic}
\end{algorithm}

\begin{table*}[th]
  \centering
\begin{adjustbox}{max width=\textwidth}
  
    \begin{tabular}{p{9.315em}cccccccccccc|c}
    \toprule
    \multicolumn{14}{c}{\textbf{(a) O-AUROC~($\uparrow$)}} \\
    \midrule
    \textbf{Method} & \multicolumn{1}{c}{\textbf{Airplane}} & \multicolumn{1}{c}{\textbf{Car}} & \multicolumn{1}{c}{\textbf{Candy}} & \multicolumn{1}{c}{\textbf{Chicken}} & \multicolumn{1}{c}{\textbf{Diamond}} & \multicolumn{1}{c}{\textbf{Duck}} & \multicolumn{1}{c}{\textbf{Fish}} & \multicolumn{1}{c}{\textbf{Gemstone}} & \multicolumn{1}{c}{\textbf{Seahorse}} & \multicolumn{1}{c}{\textbf{Shell}} & \multicolumn{1}{c}{\textbf{Starfish}} & \multicolumn{1}{c}{\textbf{Toffees}} & \multicolumn{1}{|c}{\textbf{Mean}} \\
    \midrule
    \textbf{BTF(Raw)} & 0.730 & 0.647 & 0.539 & 0.789 & 0.707 & 0.691 & 0.602 & \textcolor[rgb]{ 1,  0,  0}{\textbf{0.686}} & 0.596 & 0.396 & 0.530  & 0.703 & 0.635 \\
    \textbf{BTF(FPFH)} & 0.520  & 0.560  & 0.630  & 0.432 & 0.545 & \textcolor[rgb]{ 1,  0,  0}{\textbf{0.784}} & 0.549 & 0.648 & \textcolor[rgb]{ 1,  0,  0}{\textbf{0.779}} & \textcolor[rgb]{ 1,  0,  0}{\textbf{0.754}} & 0.575 & 0.462 & 0.603 \\
    \textbf{M3DM} & 0.434 & 0.541 & 0.552 & 0.683 & 0.602 & 0.433 & 0.540 & 0.644 & 0.495 & \textcolor[rgb]{ 0,  .439,  .753}{\textbf{0.694}} & 0.551 & 0.450 & 0.552 \\
    \textbf{PatchCore(FPFH)} & \textcolor[rgb]{ 1,  0,  0}{\textbf{0.882}} & 0.590  & 0.541 & \textcolor[rgb]{ 0,  .439,  .753}{\textbf{0.837}} & 0.574 & 0.546 & 0.675 & 0.370  & 0.505 & 0.589 & 0.441 & 0.565 & 0.593 \\
    \textbf{PatchCore(PointMAE)} & 0.726 & 0.498 & 0.663 & 0.827 & 0.783 & 0.489 & 0.630  & 0.374 & 0.539 & 0.501 & 0.519 & 0.585 & 0.594 \\
    \textbf{CPMF} & 0.701 & 0.551 & 0.552 & 0.504 & 0.523 & 0.582 & 0.558 & 0.589 & \textcolor[rgb]{ 0,  .439,  .753}{\textbf{0.729}} & 0.653 & \textcolor[rgb]{ 1,  0,  0}{\textbf{0.700}} & 0.390  & 0.586 \\
    \textbf{RegAD} & 0.716 & 0.697 & 0.685 & \textcolor[rgb]{ 1,  0,  0}{\textbf{0.852}} & 0.900   & 0.584 & \textcolor[rgb]{ 0,  .439,  .753}{\textbf{0.915}} & 0.417 & 0.762 & 0.583 & 0.506 & \textcolor[rgb]{ 0,  .439,  .753}{\textbf{0.827}} & 0.704 \\
    \textbf{IMRNet} & 0.762 & \textcolor[rgb]{ 0,  .439,  .753}{\textbf{0.711}} & \textcolor[rgb]{ 0,  .439,  .753}{\textbf{0.755}} & 0.780  & \textcolor[rgb]{ 0,  .439,  .753}{\textbf{0.905}} & 0.517 & 0.880  & \textcolor[rgb]{ 0,  .439,  .753}{\textbf{0.674}} & 0.604 & 0.665 & \textcolor[rgb]{ 0,  .439,  .753}{\textbf{0.674}} & 0.774 & \textcolor[rgb]{ 0,  .439,  .753}{\textbf{0.725}} \\
    \textbf{ISMP(Ours)} & \textcolor[rgb]{ 0,  .439,  .753}{\textbf{0.858}} & \textcolor[rgb]{ 1,  0,  0}{\textbf{0.731}} & \textcolor[rgb]{ 1,  0,  0}{\textbf{0.852}} & 0.714 & \textcolor[rgb]{ 1,  0,  0}{\textbf{0.948}} & \textcolor[rgb]{ 0,  .439,  .753}{\textbf{0.712}} & \textcolor[rgb]{ 1,  0,  0}{\textbf{0.945}} & 0.468 & \textcolor[rgb]{ 0,  .439,  .753}{\textbf{0.729}} & 0.623 & 0.660  & \textcolor[rgb]{ 1,  0,  0}{\textbf{0.842}} & \textcolor[rgb]{ 1,  0,  0}{\textbf{0.767}} \\
    \midrule\\
    \midrule
    \multicolumn{14}{c}{\textbf{(b) P-AUROC~($\uparrow$)}} \\
    \midrule
    \textbf{Method} & \multicolumn{1}{c}{\textbf{Airplane}} & \multicolumn{1}{c}{\textbf{Car}} & \multicolumn{1}{c}{\textbf{Candy}} & \multicolumn{1}{c}{\textbf{Chicken}} & \multicolumn{1}{c}{\textbf{Diamond}} & \multicolumn{1}{c}{\textbf{Duck}} & \multicolumn{1}{c}{\textbf{Fish}} & \multicolumn{1}{c}{\textbf{Gemstone}} & \multicolumn{1}{c}{\textbf{Seahorse}} & \multicolumn{1}{c}{\textbf{Shell}} & \multicolumn{1}{c}{\textbf{Starfish}} & \multicolumn{1}{c}{\textbf{Toffees}} & \multicolumn{1}{|c}{\textbf{Mean}} \\
    \midrule
    \textbf{BTF(Raw)} & 0.564 & 0.647 & 0.735 & 0.609 & 0.563 & 0.601 & 0.514 & 0.597 & 0.520  & 0.489 & 0.392 & 0.623 & 0.571 \\
    \textbf{BTF(FPFH)} & \textcolor[rgb]{ 0,  .439,  .753}{\textbf{0.738}} & 0.708 & \textcolor[rgb]{ 0,  .439,  .753}{\textbf{0.864}} & \textcolor[rgb]{ 0,  .439,  .753}{\textbf{0.735}} & \textcolor[rgb]{ 0,  .439,  .753}{\textbf{0.882}} & \textcolor[rgb]{ 0,  .439,  .753}{\textbf{0.875}} & 0.709 & \textcolor[rgb]{ 0,  .439,  .753}{\textbf{0.891}} & 0.512 & 0.571 & 0.501 & \textcolor[rgb]{ 0,  .439,  .753}{\textbf{0.815}} & \textcolor[rgb]{ 0,  .439,  .753}{\textbf{0.733}} \\
    \textbf{M3DM} & 0.547 & 0.602 & 0.679 & 0.678 & 0.608 & 0.667 & 0.606 & 0.674 & 0.560  & 0.738 & 0.532 & 0.682 & 0.631 \\
    \textbf{PatchCore(FPFH)} & 0.562 & \textcolor[rgb]{ 0,  .439,  .753}{\textbf{0.754}} & 0.780  & 0.429 & 0.828 & 0.264 & \textcolor[rgb]{ 0,  .439,  .753}{\textbf{0.829}} & \textcolor[rgb]{ 1,  0,  0}{\textbf{0.910}} & 0.739 & 0.739 & 0.606 & 0.747 & 0.682 \\
    \textbf{PatchCore(PointMAE)} & 0.569 & 0.609 & 0.627 & 0.729 & 0.718 & 0.528 & 0.717 & 0.444 & 0.633 & 0.709 & 0.580  & 0.580  & 0.620\\
    \textbf{RegAD} & 0.631 & 0.718 & 0.724 & 0.676 & 0.835 & 0.503 & 0.826 & 0.545 & \textcolor[rgb]{ 1,  0,  0}{\textbf{0.817}} & \textcolor[rgb]{ 0,  .439,  .753}{\textbf{0.811}} & \textcolor[rgb]{ 0,  .439,  .753}{\textbf{0.617}} & 0.759 & 0.705 \\
    \textbf{ISMP(Ours)} & \textcolor[rgb]{ 1,  0,  0}{\textbf{0.753}} & \textcolor[rgb]{ 1,  0,  0}{\textbf{0.836}} & \textcolor[rgb]{ 1,  0,  0}{\textbf{0.907}} & \textcolor[rgb]{ 1,  0,  0}{\textbf{0.798}} & \textcolor[rgb]{ 1,  0,  0}{\textbf{0.926}} & \textcolor[rgb]{ 1,  0,  0}{\textbf{0.876}} & \textcolor[rgb]{ 1,  0,  0}{\textbf{0.886}} & 0.857 & \textcolor[rgb]{ 0,  .439,  .753}{\textbf{0.813}} & \textcolor[rgb]{ 1,  0,  0}{\textbf{0.839}} & \textcolor[rgb]{ 1,  0,  0}{\textbf{0.641}} & \textcolor[rgb]{ 1,  0,  0}{\textbf{0.895}} & \textcolor[rgb]{ 1,  0,  0}{\textbf{0.836}} \\
    \bottomrule
    \end{tabular}%
 \end{adjustbox}
\caption{The experimental results of (a) O-AUROC~($\uparrow$) and (b) P-AUROC~($\uparrow$) for anomaly detection across 12 categories of Real3D-AD. The best and the second-best results are highlighted in \textcolor[rgb]{1, 0, 0}{\textbf{red}}  and \textcolor[rgb]{0, .439, .753}{\textbf{blue}}, respectively. Our model achieved the best average performance across the 12 categories for both metrics.}
  \label{xlx1}%
\end{table*}%

\begin{table*}[th]
  \centering
  \begin{adjustbox}{max width=\textwidth}
    \begin{tabular}{l|*{14}{S[table-format=1.3]}}
    \toprule
    \multicolumn{15}{c}{\textbf{P-AUROC~($\uparrow$)}} \\
    \midrule
    \textbf{Method} & \multicolumn{1}{c}{\textbf{cap0}} & \multicolumn{1}{c}{\textbf{cap3}} & \multicolumn{1}{c}{\textbf{helmet3}} & \multicolumn{1}{c}{\textbf{cup0}} & \multicolumn{1}{c}{\textbf{bowl4}} & \multicolumn{1}{c}{\textbf{vase3}} & \multicolumn{1}{c}{\textbf{headset1}} & \multicolumn{1}{c}{\textbf{eraser0}} & \multicolumn{1}{c}{\textbf{vase8}} & \multicolumn{1}{c}{\textbf{cap4}} & \multicolumn{1}{c}{\textbf{vase2}} & \multicolumn{1}{c}{\textbf{vase4}} & \multicolumn{1}{c}{\textbf{helmet0}} & \multicolumn{1}{c}{\textbf{bucket1}} \\
    \midrule
    \textbf{BTF(Raw)} & 0.524 & 0.687 & 0.700   & 0.632 & 0.563 & 0.602 & 0.475 & 0.637 & 0.550 & 0.469 & 0.403 & 0.613 & 0.504 & 0.686 \\
    \textbf{BTF(FPFH)} & \textcolor[rgb]{ 0,  .439,  .753}{\textbf{0.730}} & 0.658 & \textcolor[rgb]{ 0,  .439,  .753}{\textbf{0.724}} & \textcolor[rgb]{ 0,  .439,  .753}{\textbf{0.790}} & 0.679 & \textcolor[rgb]{ 0,  .439,  .753}{\textbf{0.699}} & 0.591 & \textcolor[rgb]{ 0,  .439,  .753}{\textbf{0.719}} & 0.662 & 0.524 & 0.646 & 0.710 & 0.575 & 0.633 \\
    \textbf{M3DM} & 0.531 & 0.605 & 0.655 & 0.715 & 0.624 & 0.658 & 0.585 & 0.710  & 0.551 & 0.718 & \textcolor[rgb]{ 0,  .439,  .753}{\textbf{0.737}} & \textcolor[rgb]{ 0,  .439,  .753}{\textbf{0.655}} & 0.599 & 0.699 \\
    \textbf{Patchcore(FPFH)} & 0.472 & 0.653 & \textcolor[rgb]{ 1,  0,  0}{\textbf{0.737}} & 0.655 & \textcolor[rgb]{ 0,  .439,  .753}{\textbf{0.720}} & 0.430 & 0.464 & 0.810 & 0.575 & 0.595 & 0.721 & 0.505 & 0.548 & 0.571 \\
    \textbf{Patchcore(PointMAE)} & 0.544 & 0.488 & 0.615 & 0.510 & 0.501 & 0.465 & 0.423 & 0.378 & 0.364 & 0.725 & \textcolor[rgb]{ 1,  0,  0}{\textbf{0.742}} & 0.523 & 0.580 & \textcolor[rgb]{ 0,  .439,  .753}{\textbf{0.754}} \\
    \textbf{CPMF} & 0.601 & 0.551 & 0.520 & 0.497 & 0.683 & 0.582 & 0.458 & 0.689 & 0.529 & 0.553 & 0.582 & 0.514 & 0.555 & 0.601 \\
    \textbf{RegAD} & 0.632 & \textcolor[rgb]{ 0,  .439,  .753}{\textbf{0.718}} & 0.620 & 0.685 & 0.800 & 0.511 & \textcolor[rgb]{ 0,  .439,  .753}{\textbf{0.626}} & \textcolor[rgb]{ 1,  0,  0}{\textbf{0.755}} & \textcolor[rgb]{ 0,  .439,  .753}{\textbf{0.811}} & \textcolor[rgb]{ 1,  0,  0}{\textbf{0.815}} & 0.405 & \textcolor[rgb]{ 1,  0,  0}{\textbf{0.755}} & \textcolor[rgb]{ 0,  .439,  .753}{\textbf{0.600}} & 0.725 \\
    \textbf{IMRNet} & 0.715 & 0.706 & 0.663 & 0.643 & 0.576 & 0.401 & 0.476 & 0.548 & 0.635 & \textcolor[rgb]{ 0,  .439,  .753}{\textbf{0.753}} & 0.614 & 0.524 & 0.598 & \textcolor[rgb]{ 1,  0,  0}{\textbf{0.774}} \\
    \textbf{ISMP(Ours)} & \textcolor[rgb]{ 1,  0,  0}{\textbf{0.865}} & \textcolor[rgb]{ 1,  0,  0}{\textbf{0.734}} & 0.722 & \textcolor[rgb]{ 1,  0,  0}{\textbf{0.869}} & \textcolor[rgb]{ 1,  0,  0}{\textbf{0.740}} & \textcolor[rgb]{ 1,  0,  0}{\textbf{0.762}} & \textcolor[rgb]{ 1,  0,  0}{\textbf{0.702}} & 0.706 & \textcolor[rgb]{ 1,  0,  0}{\textbf{0.851}} & \textcolor[rgb]{ 0,  .439,  .753}{\textbf{0.753}} & 0.733 & 0.545 & \textcolor[rgb]{ 1,  0,  0}{\textbf{0.683}} & 0.672 \\
    \midrule
    \multicolumn{1}{c}{} &       &       &       &       &       &       &       &       &       &       &       &       &       &  \\
    \midrule
    \textbf{Method} & \multicolumn{1}{c}{\textbf{bottle3}} & \multicolumn{1}{c}{\textbf{vase0}} & \multicolumn{1}{c}{\textbf{bottle0}} & \multicolumn{1}{c}{\textbf{tap1}} & \multicolumn{1}{c}{\textbf{bowl0}} & \multicolumn{1}{c}{\textbf{bucket0}} & \multicolumn{1}{c}{\textbf{vase5}} & \multicolumn{1}{c}{\textbf{vase1}} & \multicolumn{1}{c}{\textbf{vase9}} & \multicolumn{1}{c}{\textbf{ashtray0}} & \multicolumn{1}{c}{\textbf{bottle1}} & \multicolumn{1}{c}{\textbf{tap0}} & \multicolumn{1}{c}{\textbf{phone}} & \multicolumn{1}{c}{\textbf{cup1}} \\
    \midrule
    \textbf{BTF(Raw)} & 0.720 & 0.618 & 0.551 & 0.564 & 0.524 & 0.617 & 0.585 & 0.549 & 0.564 & 0.512 & 0.491 & 0.527 & 0.583 & 0.561 \\
    \textbf{BTF(FPFH)} & 0.622 & 0.642 & 0.641 & 0.596 & 0.710 & 0.401 & 0.429 & 0.619 & 0.568 & 0.624 & 0.549 & 0.568 & 0.675 & 0.619 \\
    \textbf{M3DM} & 0.532 & 0.608 & 0.663 & 0.712 & 0.658 & \textcolor[rgb]{ 1,  0,  0}{\textbf{0.698}} & 0.642 & 0.602 & 0.663 & 0.577 & 0.637 & 0.654 & 0.358 & 0.556 \\
    \textbf{Patchcore(FPFH)} & 0.512 & 0.655 & 0.654 & \textcolor[rgb]{ 1,  0,  0}{\textbf{0.768}} & 0.524 & 0.459 & 0.447 & 0.453 & 0.663 & 0.597 & 0.687 & \textcolor[rgb]{ 0,  .439,  .753}{\textbf{0.733}} & 0.488 & 0.596 \\
    \textbf{Patchcore(PointMAE)} & \textcolor[rgb]{ 0,  .439,  .753}{\textbf{0.653}} & \textcolor[rgb]{ 1,  0,  0}{\textbf{0.677}} & 0.553 & 0.541 & 0.527 & 0.586 & 0.572 & 0.551 & 0.423 & 0.495 & 0.606 & \textcolor[rgb]{ 1,  0,  0}{\textbf{0.858}} & \textcolor[rgb]{ 1,  0,  0}{\textbf{0.886}} & \textcolor[rgb]{ 1,  0,  0}{\textbf{0.856}} \\
    \textbf{CPMF} & 0.435 & 0.458 & 0.521 & 0.657 & 0.745 & 0.486 & \textcolor[rgb]{ 0,  .439,  .753}{\textbf{0.651}} & 0.486 & 0.545 & 0.615 & 0.571 & 0.458 & 0.545 & 0.509 \\
    \textbf{RegAD} & 0.525 & 0.548 & \textcolor[rgb]{ 1,  0,  0}{\textbf{0.888}} & \textcolor[rgb]{ 0,  .439,  .753}{\textbf{0.741}} & 0.775 & \textcolor[rgb]{ 0,  .439,  .753}{\textbf{0.619}} & 0.624 & 0.602 & \textcolor[rgb]{ 1,  0,  0}{\textbf{0.694}} & \textcolor[rgb]{ 1,  0,  0}{\textbf{0.698}} & \textcolor[rgb]{ 0,  .439,  .753}{\textbf{0.696}} & 0.589 & 0.599 & \textcolor[rgb]{ 0,  .439,  .753}{\textbf{0.698}} \\
    \textbf{IMRNet} & 0.641 & 0.535 & 0.556 & 0.699 & \textcolor[rgb]{ 0,  .439,  .753}{\textbf{0.781}} & 0.585 & \textcolor[rgb]{ 1,  0,  0}{\textbf{0.682}} & \textcolor[rgb]{ 0,  .439,  .753}{\textbf{0.685}} & \textcolor[rgb]{ 0,  .439,  .753}{\textbf{0.691}} & \textcolor[rgb]{ 0,  .439,  .753}{\textbf{0.671}} & \textcolor[rgb]{ 1,  0,  0}{\textbf{0.702}} & 0.681 & \textcolor[rgb]{ 0,  .439,  .753}{\textbf{0.742}} & 0.688 \\
    \textbf{ISMP(Ours)} & \textcolor[rgb]{ 1,  0,  0}{\textbf{0.775}} & \textcolor[rgb]{ 0,  .439,  .753}{\textbf{0.661}} & \textcolor[rgb]{ 0,  .439,  .753}{\textbf{0.770}} & 0.552 & \textcolor[rgb]{ 1,  0,  0}{\textbf{0.851}} & 0.524 & 0.472 & \textcolor[rgb]{ 1,  0,  0}{\textbf{0.843}} & 0.615 & 0.603 & 0.568 & 0.522 & 0.661 & 0.600 \\
    \midrule
    \multicolumn{1}{c}{} & \textcolor[rgb]{ 1,  0,  0}{} &       &       &       &       &       &       &       &       &       &       &       &       &  \\
\cmidrule{1-14}
    \textbf{Method} & \multicolumn{1}{c}{\textbf{vase7}} & \multicolumn{1}{c}{\textbf{helmet2}} & \multicolumn{1}{c}{\textbf{cap5}} & \multicolumn{1}{c}{\textbf{shelf0}} & \multicolumn{1}{c}{\textbf{bowl5}} & \multicolumn{1}{c}{\textbf{bowl3}} & \multicolumn{1}{c}{\textbf{helmet1}} & \multicolumn{1}{c}{\textbf{bowl1}} & \multicolumn{1}{c}{\textbf{headset0}} & \multicolumn{1}{c}{\textbf{bag0}} & \multicolumn{1}{c}{\textbf{bowl2}} & \multicolumn{1}{c|}{\textbf{jar0}} & \multicolumn{1}{|c}{\textbf{Mean}} \\
\cmidrule{1-14}
    \textbf{BTF(Raw)} & 0.578 & 0.605 & 0.373 & 0.464 & 0.517 & 0.685 & 0.449 & 0.464 & 0.578 & 0.430  & 0.426 & 0.423 & \multicolumn{1}{|c}{0.550} \\
    \textbf{BTF(FPFH)} & 0.540  & 0.643 & 0.586 & 0.619 & 0.699 & \textcolor[rgb]{0, .439, .753}{\textbf{0.690}} & \textcolor[rgb]{1, 0, 0}{\textbf{0.749}} & \textcolor[rgb]{1, 0, 0}{\textbf{0.768}} & 0.620  & \textcolor[rgb]{0, .439, .753}{\textbf{0.746}} & 0.518 & 0.427 & \multicolumn{1}{|c}{0.628} \\
    \textbf{M3DM} & 0.517 & 0.623 & 0.655 & 0.554 & 0.489 & 0.657 & 0.427 & 0.663 & 0.581 & 0.637 & \textcolor[rgb]{0, .439, .753}{\textbf{0.694}} & 0.541 & \multicolumn{1}{|c}{0.616} \\
    \textbf{Patchcore(FPFH)} & 0.693 & 0.455 & \textcolor[rgb]{1, 0, 0}{\textbf{0.795}} & 0.613 & 0.358 & 0.327 & 0.489 & 0.531 & 0.583 & 0.574 & 0.625 & 0.478 & \multicolumn{1}{|c}{0.580} \\
    \textbf{Patchcore(PointMAE)} & 0.651 & 0.651 & 0.545 & 0.543 & 0.562 & 0.581 & 0.562 & 0.524 & 0.575 & 0.674 & 0.515 & 0.487 & \multicolumn{1}{|c}{0.577} \\
    \textbf{CPMF} & 0.504 & 0.515 & 0.551 & \textcolor[rgb]{1, 0, 0}{\textbf{0.783}} & 0.684 & 0.641 & 0.542 & 0.488 & \textcolor[rgb]{1, 0, 0}{\textbf{0.699}} & 0.655 & 0.635 & 0.611 & \multicolumn{1}{|c}{0.573} \\
    \textbf{RegAD} & \textcolor[rgb]{1, 0, 0}{\textbf{0.881}} & 0.825 & 0.467 & 0.688 & \textcolor[rgb]{0, .439, .753}{\textbf{0.691}} & 0.654 & \textcolor[rgb]{0, .439, .753}{\textbf{0.624}} & 0.645 & 0.580 & 0.715 & 0.593 & 0.599 & \multicolumn{1}{|c}{\textcolor[rgb]{0, .439, .753}{\textbf{0.668}}} \\
    \textbf{IMRNet} & 0.593 & 0.644 & \textcolor[rgb]{0, .439, .753}{\textbf{0.742}} & 0.605 & \textcolor[rgb]{1, 0, 0}{\textbf{0.715}} & 0.599 & 0.604 & \textcolor[rgb]{0, .439, .753}{\textbf{0.705}} & 0.615 & 0.668 & 0.684 & \textcolor[rgb]{0, .439, .753}{\textbf{0.765}} & \multicolumn{1}{|c}{0.650} \\
    \textbf{ISMP(Ours)} & \textcolor[rgb]{0, .439, .753}{\textbf{0.701}} & \textcolor[rgb]{1, 0, 0}{\textbf{0.844}} & 0.678 & \textcolor[rgb]{0, .439, .753}{\textbf{0.687}} & 0.534 & \textcolor[rgb]{1, 0, 0}{\textbf{0.773}} & 0.622 & 0.546 & 0.580 & \textcolor[rgb]{1, 0, 0}{\textbf{0.747}} & \textcolor[rgb]{1, 0, 0}{\textbf{0.736}} & \textcolor[rgb]{1, 0, 0}{\textbf{0.823}} & \multicolumn{1}{|c}{\textcolor[rgb]{1, 0, 0}{\textbf{0.691}}} \\
\cmidrule{1-14}
    \end{tabular}%
  \end{adjustbox}
  \caption{The experimental results of P-AUROC~($\uparrow$) for anomaly detection of 40 categories of Anomaly-ShapeNet. The best and the second-best results are highlighted in \textcolor[rgb]{1, 0, 0}{\textbf{red}} and \textcolor[rgb]{0, .439, .753}{\textbf{blue}}, respectively. Our model achieved better performance in pixel-level anomaly detection. 
  }
  \label{xlx2}%
\end{table*}%

\subsection{Anomaly Score Calculation}
We utilize the feature memory bank $\mathcal{M}^C$ and the coordinate memory bank $\mathcal{M}^F$ to compute anomaly scores. Here, we illustrate the scoring process using the $\mathcal{M}^F$ memory bank as an example. We find the nearest neighbor in the $\mathcal{M}^F$ for the test object's point-level feature $\mathcal{P}(m^{\text{test}})$. The nearest neighbor search method~\cite{liu2023real3daddatasetpointcloud} is denoted as:
\begin{equation}
\begin{aligned}
&m^{\text{test},*} = \arg \max_{m^{\text{test}} \in \mathcal{P}(x^{\text{test}})} \min_{m' \in \mathcal{M}^F} \| m^{\text{test}} - m' \|_2, \\
&m^F_* = \arg \min_{m' \in \mathcal{M}^F} \| m^{\text{test}} - m' \|_2.
\end{aligned}
\end{equation}
Calculate the nearest neighbor distance as the local feature anomaly score $s^F_*$:
\begin{equation}
s^F_* = \| m^{\text{test},*} - m^F_* \|_2.
\end{equation}
The anomaly score is adjusted using a re-weighting method \cite{Liu_2016}, which is denoted as:
\begin{equation}
s^F = \left( 1 - \frac{\exp \| m^{\text{test},*} - m^F_* \|_2}{
\sum_{m \in N_3(m^*)} \exp \| m^{\text{test},*} - m \|_2} \right) s^F_*.
\end{equation}
Here, $N_3(m^*)$ represents the 3 nearest features in the $\mathcal{M}^F$. Perform similar calculations using the $\mathcal{M}^C$ to obtain the coordinate anomaly score $s^C$. 
\begin{equation}\label{eq:st}
s = \frac{s^F + s^C}{2}.
\end{equation}
Compute the overall anomaly score for each point cloud $s$ by averaging the $s^F$ and $s^C$ using Equ.~\eqref{eq:st}.

\section{Experiments}
In this section, we firstly evaluated the effectiveness of ISMP in the anomaly detection task and secondly supplemented the assessment with the generalization ability of SIE across multiple tasks.


\subsection{Implementation}

\textbf{Datasets.} We conducted comparative experiments on two mainstreaming datasets, namely Real3D-AD and Anomaly-ShapeNet.
(1) The Real3D-AD dataset ~\cite{liu2023real3daddatasetpointcloud} is a high-resolution, large-scale anomaly dataset containing 1,254 samples across 12 categories. The training set for each category includes four normal samples, while the test set for each category contains both normal samples and anomalous samples with various defects. (2) The Anomaly-ShapeNet dataset ~\cite{li2023scalable3danomalydetection} provides 40 categories, containing over 1,600 positive and negative samples. The training set for each category includes four normal samples, while the test set for each category contains both normal samples and anomalous samples with various defects.

\textbf{Baselines.}
We selected BTF~\cite{horwitz2022featureclassical3dfeatures},  M3DM~\cite{wang2023multimodalindustrialanomalydetection}, PatchCore~\cite{roth2022totalrecallindustrialanomaly}, CPMF~\cite{cao2023complementarypseudomultimodalfeature}, RegAD~\cite{liu2023real3daddatasetpointcloud}, and IMRNet~\cite{li2023scalable3danomalydetection} for comparison. Note that BTF(FPFH) denotes that we incorporate fast point feature histogram~\cite{5152473}. The results of these methods are obtained through publicly available code or referenced papers.

\textbf{Evaluation Metrics.} For the anomaly detection task, we use P-AUROC ($\uparrow$) to evaluate pixel-level anomaly localization capability and O-AUROC ($\uparrow$) to evaluate object-level anomaly detection capability. Higher values for both metrics indicate a more robust anomaly detection capability. 

\textbf{Experimental Details.}
The experiments are conducted on a machine equipped with an RTX 3090 (24GB) GPU. For ISMP, we used the pre-trained weights of the PointMAE and EfficientNet~\cite{tan2020efficientnetrethinkingmodelscaling} models to complete our experiments. Our parameter settings for ISMP followed those of RegAD~\cite{liu2023real3daddatasetpointcloud}, with $\alpha$ set to 0.2, $\beta$ set to 0.2, and $\gamma$ set to 0.001.

\subsection{Main Results}

\textbf{Comparisons on Real3D-AD.} The quantitative comparisons of ISMP with competing models are presented in Table~\ref{xlx1}. We observed that other models exhibit biases in high-precision anomaly localization tasks, making accurate localization challenging. Our method achieved O-AUROC and P-AUROC scores of 0.757 and 0.836, respectively, significantly improving state-of-the-art (SOTA) methods.

\textbf{Comparisons on Anomaly-ShapeNet.} In Table~\ref{xlx2}, we quantitatively analyze the pixel-level anomaly detection results on the Anomaly-ShapeNet dataset. Due to the diversity of the training set, Anomaly-ShapeNet presents challenges for better utilization of features. Our method achieved a P-AUROC score of 0.691, outperforming previous methods.

\subsection{Ablation Study}

\textbf{Evaluation of the ISMP Efficiency.}
We evaluated each module's effectiveness on Real3D-AD, as summarized in Table~\ref{ablatio_n}. The worst performance occurred with only coordinates and PointMAE features, emphasizing the need for improved local coordinate representation. Incorporating PFPH around sampled points increased P-AUROC by 18.9\%, and further optimization with a feature filtering module added another 1.8\%. Without global features, O-AUROC stayed at 65.6\%, but introducing internal spatial modality features raised it by 10.1\%. Notably, using only the two outer projection slices, omitting the internal slice, produced the second-best results, showing internal features are more reliable. These findings confirm our model's optimal composition.

\begin{table}[ht]
  \centering
  \begin{tabular}{lcc}
    \toprule
    \textbf{Method} & \textbf{I-AUROC~($\uparrow$)} & \textbf{P-AUROC~($\uparrow$)} \\
    \midrule
    \textbf{ISMP} & \textbf{0.757} & \textbf{0.836} \\
    ISMP$_{w/o\ I}$ & 0.656 & 0.827 \\
    ISMP$_{w/o\ I\&F}$ & 0.623 & 0.809 \\
    ISMP$_{w/o\ I, F\&E}$ & 0.594 & 0.620 \\
    ISMP$_{O}$ & 0.717 &0.812 \\
    \midrule
  \end{tabular}
 \caption{Ablation study results. $I$ and $F$ represent SIE and the filter module respectively. $E$ stands for enhanced feature extraction. $O$ stands for using only the external projection and eliminating the internal projection. }
   \label{ablatio_n}
\end{table}
\begin{table}[ht]
  \centering
    \begin{tabular}{p{2.7cm}|p{1.5cm}<\centering p{1.5cm}<\centering}
    \toprule
    \textbf{Method} & \multicolumn{1}{c}{\textbf{Training}} & \multicolumn{1}{c}{\textbf{Test}} \\
    \midrule
    RegAD & 5.52  & 7.71 \\
    M3DM & 4.50  & 6.43 \\
    \textbf{ISMP (Ours)} & \textbf{2.78} & \textbf{4.37} \\
    \bottomrule
    \end{tabular}%
  \caption{The average training and test time in terms of FPS ($\uparrow$, per sample) for models evaluated on the \textit{seahorse} class.}
  \label{cost}%
\end{table}%

\textbf{Evaluation of the ISMP Effectivness.}
Our model achieved outstanding performance but grappled with inference efficiency challenges linked to incorporating extra modality information, as evidenced in Table~\ref{cost}. The ISMP lacks sufficient training and reasoning speed compared to its competitors.

\begin{figure}[h] 
    \centering 
    \includegraphics[width=1\columnwidth]{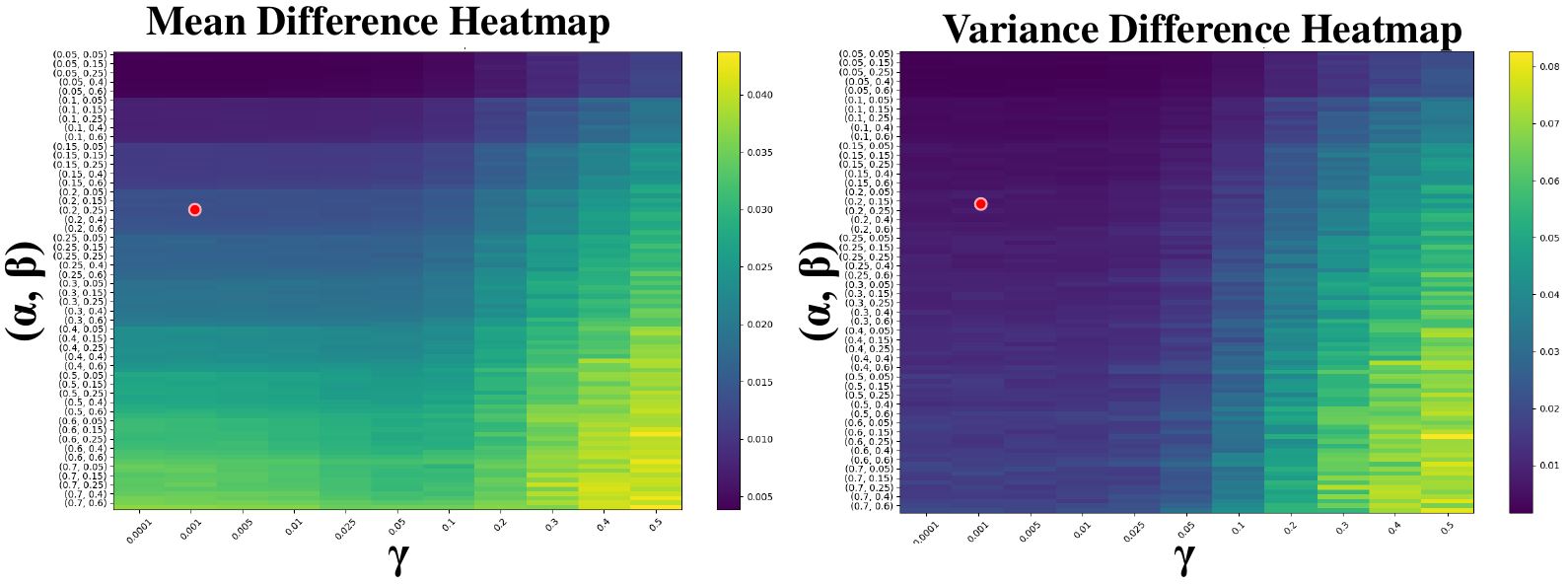} 
    \caption{Heatmaps of the impact of parameters $\alpha$, $\beta$, $\gamma$ on \texttt{mean} and \texttt{variance}. The ordinate represents a combination of $\alpha$ and $\beta$, and the abscissa represents $\gamma$. The lighter the color of the block in the figure, the larger the difference before and after the transformation. The red marks in the figure are the parameters we selected.}
    \label{fig:heat} 
\end{figure}

\textbf{Analysis of the Feature Filtering Module.}
The mean and variance of feature matrices play a crucial role in anomaly detection, as more concentrated feature distributions are beneficial for detecting anomalies ~\cite{wang2023multimodalindustrialanomalydetection}. To further investigate the impact of the feature filtering module on feature matrices under different parameter settings, we present the effects of various parameters on the feature matrices in Figure~\ref{fig:heat}. The variance of the feature matrix has a greater impact on distinguishing between abnormal and normal features than the mean. Specifically, controlling the variance in the feature matrix is crucial for better feature distinction. Notably, features from PointMAE resemble a standard normal distribution after normalization. We randomly selected one thousand feature matrices conforming to this distribution for feature filtering and analyzed their effects on the mean and variance. In practice, we conservatively chose the parameters labeled in the diagram. Our proposed feature filtering module ultimately exhibits controllability over the features, resulting in improved feature representations.

\subsection{Evaluation of SIE Generalization}
To verify the effectiveness of SIE for the feature perception in 3D anomaly detection, we have designed two related tasks for point clouds, namely classification and segmentation. From the results, we can observe that SIE can provide more adequate information.

\textbf{Datasets.} 
(1) The ModelNet40 is a point cloud classification dataset containing forty categories~\cite{wu20153dshapenetsdeeprepresentation}. We use it to test the effect of SIE on classification tasks to prove generalization. 
(2) The ShapeNet-Part is a point cloud dataset commonly used for semantic segmentation. It includes ten categories of standard household items, with many 3D models corresponding to each category~\cite{chang2015shapenetinformationrich3dmodel}. We use it to test the impact of SIE on semantic segmentation tasks to prove generalization.
    
\textbf{Evaluation Metrics.}
We use Accuracy ($\uparrow$) to evaluate the model's capability for the point cloud classification task. For the segmentation task, we use Instance average Intersection over Union~(IoU, $\uparrow$) to assess the model's adaptability.

\textbf{Baselines.}
To demonstrate its significant role in extracting global information from point clouds, we analyzed the effect of SIE in point cloud classification on ModelNet40, comparing it with Subvolume~\cite{qi2016volumetricmultiviewcnnsobject}, MVCNN~\cite{7410471}, PointNet~\cite{qi2017pointnetdeeplearningpoint}, and PointNet++~\cite{qi2017pointnetdeephierarchicalfeature}. Moreover, we tested the role of SIE in semantic segmentation on ShapeNet-Part, comparing it with Yi~\cite{10.1145/2980179.2980238}, PointNet, SSCNN~\cite{8100180} and PointNet++. The results of these methods are obtained through publicly available code or referenced papers. We employed the same PointNet++ settings as those used in~\cite{qi2017pointnetdeephierarchicalfeature}.

\textbf{Comparison Results on ModelNet40.}
We used intra-space pseudo-modality as a crucial supplementary input for point cloud classification in PointNet++. As shown in Table~\ref{Model401}, a simple feature injection is sufficient to enhance the performance of point cloud classification, since the intra-space pseudo-modality provides significant additional information for point clouds. It is proved that SIE has potential in point cloud classification tasks.

\begin{table}[htbp]
  \centering
    \begin{tabular}{p{3cm}|p{2cm}<\centering p{2cm}<\centering}
    \toprule
    \textbf{Method} & \textbf{Input} & \multicolumn{1}{c}{\textbf{Accuracy~(\%)}} \\
    \midrule
    Subvolume & vox   & 89.2 \\
    MVCNN & img   & 90.1 \\
    PointNet(vanilla) & pc    & 87.2 \\
    PointNet & pc    & 89.2 \\
    PointNet++ & pc    & 90.7 \\
    PointNet++(SIE) & pc    & \textbf{91.1} \\
    \midrule
    \end{tabular}%
  \caption{Shape classification results on ModelNet40.}
    \label{Model401}%
\end{table}

\textbf{Comparison Results on ShapeNet-Part.}
Using the global information from SIE as a supplement to features extracted by PointNet++ for part segmentation, we observed enhanced performance, as shown in Table ~\ref{Shapenetpart}. The results demonstrate SIE's potential in aligning local and global information. Overall, ISMP excels in 3D anomaly detection, while SIE shows strong generalization and robustness, making it adaptable to other tasks.
\begin{table}[ht]
  \centering
    \begin{tabular}{p{7em}|p{1.5cm}<\centering}
    \toprule
    \textbf{Model} & \multicolumn{1}{c}{\textbf{IoU~(\%)}} \\
    \midrule
    Yi    & 81.4 \\
    PointNet    & 83.7 \\
    SSCNN & 84.7 \\
    PointNet++ & 85.1 \\
    PointNet++(SIE) & \textbf{85.4} \\
    \bottomrule
    \end{tabular}%
  \caption{Segmentation results on ShapeNet-Part.}
\label{Shapenetpart}
\end{table}%

\section{Conclusion}
We propose a novel 3D AD method equipped with Internal Spatial Modality Perception~(ISMP) to address the issue of underutilizing internal information in samples. 
Our approach consists of three modules, namely a novel perception module based on the Spatial Insight Engine~(SIE), an enhanced feature extraction module, and a feature filtering module.
The experimental results demonstrate the effectiveness of our proposed method.
Besides, we verified the effectiveness of ISMP in the AD task and the generalization ability of SIE.
\textbf{Limitation.} Given the limits of the test cost, we aim to improve the model's inference speed in future work.

 \section{Acknowledgments}
This work was supported by the National Natural Science Foundation of China (Grant Nos. 62206122, 62476171, 82261138629, 62302309), the Guangdong Basic and Applied Basic Research Foundation (No. 2024A1515011367), the Guangdong Provincial Key Laboratory (Grant No. 2023B1212060076), Shenzhen Municipal Science and Technology Innovation Council (Grant No. JCYJ20220531101412030), Tencent ``Rhinoceros Birds” - Scientific Research Foundation for Young Teachers of Shenzhen University, and the Internal Fund of National Engineering Laboratory for Big Data System Computing Technology (Grant No. SZU-BDSC-IF2024-08).

\bibliography{aaai25}

\end{document}